%% file: main.tex
\pdfoutput=1

\documentclass[11pt]{article}

\usepackage{EACL2023}
\usepackage{times}
\usepackage{latexsym}
\usepackage{multirow}
\usepackage[T1]{fontenc}
\usepackage[utf8]{inputenc}
\usepackage{microtype}
\usepackage{inconsolata}
\usepackage{float}
\usepackage{graphicx}
\usepackage{booktabs}
\usepackage{amsmath}

\newcommand{\company}{\emph{Allegro}}
\newcommand{\methodname}{\emph{Conv}}
\newcommand{\banking}{\emph{Banking77}}
\newcommand{\clinc}{\emph{CLINC150}}
\newcommand{\purchase}{\emph{Purchase}}
\newcommand{\delivery}{\emph{Delivery}}
\newcommand{\retail}{\emph{Retail}}

\newcommand{\cxweakly}{\emph{CX weakly supervised}}

\newcommand{\tabbanking}{\bf{\emph{Banking77}}}
\newcommand{\tabclinc}{\bf{\emph{CLINC150}}}
\newcommand{\tabpurchase}{\bf{\emph{Purchase}}}
\newcommand{\tabdelivery}{\bf{\emph{Delivery}}}
\newcommand{\tabretail}{\bf{\emph{Retail}}}

\newcommand{\herbert}{\emph{HerBERT}}
\newcommand{\mailbert}{\emph{AlleBERT}}  
\newcommand{\mailbertconvert}{\emph{AlleConveRT}}
\newcommand{\mailberttags}{\emph{TagBERT}}
\newcommand{\bertbase}{\emph{BERT-base}}
\newcommand{\tabherbert}{\herbert}
\newcommand{\tabmailbert}{\mailbert}
\newcommand{\tabmailbertconvert}{\mailbertconvert}
\newcommand{\tabmailberttags}{\mailberttags}

\newcommand*\rot{\rotatebox{90}}
\newcommand{\linebreakcell}[2][c]{%
  \begin{tabular}[#1]{@{}c@{}}#2\end{tabular}}

\title{Going beyond research datasets: Novel intent discovery in the industry setting}

\author{Aleksandra Chrabrowa,
        Tsimur Hadeliya,
        Dariusz Kajtoch,\\
        {\bf Robert Mroczkowski},
        {\bf Piotr Rybak} \\
        ML Research at Allegro, Poznań, Poland \\
        \texttt{\{firstname.lastname\}@allegro.pl}}

\begin{document}
\maketitle
\begin{abstract}
Novel intent discovery automates the process of grouping similar messages (questions) to identify previously unknown intents. However, current research focuses on publicly available datasets which have only the question field and significantly differ from real-life datasets. This paper proposes methods to improve the intent discovery pipeline deployed in a large e-commerce platform. We show the benefit of pre-training language models on in-domain data: both self-supervised and with weak supervision. We also devise the best method to utilize the conversational structure (i.e., question and answer) of real-life datasets during fine-tuning for clustering tasks, which we call \methodname. All our methods combined to fully utilize real-life datasets give up to 33pp performance boost over state-of-the-art Constrained Deep Adaptive Clustering (CDAC)~\citep{CDAC} model for question only. By comparison CDAC model for the question data only gives only up to 13pp performance boost over the naive baseline.
\end{abstract}

\section{Introduction}
\company\ is one of largest the e-commerce marketplace in Central Eastern Europe region that connects buyers and merchants. It has millions of active users. Therefore, the good functioning of the Customer Experience (CX) department is crucial as it provides the necessary support, resolves emerging issues, and answers user questions.

Task-oriented chatbots relieve humans by automatically resolving the most repetitive and trivial issues. They usually have a pre-defined set of user intents with matching template answers. Then, when a user asks a question, the intent classifier detects the question intent and returns the matching response.
Creating a reliable and comprehensive chatbot requires massive work to discover, define, and maintain a set of intents with training examples. 
With the continuous development of marketplace platforms, new intents constantly appear as new features are introduced. Therefore, the automated intent discovery system becomes a critical component.

Novel intent discovery is performed offline on historical data. In the context of personalized intelligence assistants existing approaches~\citep{CDAC,graphire2021,advin2022} focus on learning transferable features with utterance encoders that guide the discovery on unlabeled data with a handful of labeled examples belonging to known intents. However, at \company\, our main communication form is emails, and 
we have access to much richer conversational data that can improve discovery performance. A large body of historical conversational data (user questions and consultants' answers) can be leveraged in two ways. Firstly, to better initialize message encoders and secondly by performing intent discovery on conversational data as an additional signal. Additionally, a form of weak supervision is available: keywords (or tags) added by the consultants that help them understand past cases.

The paper's main contribution is the demonstration that incorporating additional signals like conversational structure or weak labels into the existing intent discovery method results in better overall performance. We pre-trained for domain adaptation three encoders using conversational data and weak labels. We devised \methodname, a method for fine-tuning on conversational data (i.e., question and answer) for the clustering task using a three-headed encoder. To the best of our knowledge, this result was not reported in the public literature. 

\section{Related Work}
\subsection{Discovering novel intents}
The goal of novel intent discovery is to identify groups of similar utterances in unlabeled data with the assistance of limited labeled data. The Constrained Deep Adaptive Clustering~\citep[CDAC]{CDAC} uses dense intent representation on top of the pre-trained BERT backbone to learn similarity functions in a semi-supervised contrastive manner. It is then utilized in the clustering algorithm. In a real-world scenario of personal assistants~\citep{graphire2021, advin2022} use a pre-trained BERT model as a backbone encoder with supervised contrastive learning to transfer distance function to unlabeled data for clustering. Unlike this work, the authors use only the question field and English \bertbase\  uncased model for initialization. They do not use in-domain unlabeled data or weak supervision for backbone pre-training.

\subsection{Transfer learning}
General-purpose pre-trained encoders like BERT are not ideal. Tasks involving domain-specific texts like, e.g., science corpus, clinical notes, or e-commerce product descriptions benefit more from additional pre-training on in-domain data due to better suited vocabulary and word embeddings to domain specific problems~\citep{beltagy-etal-2019-scibert,clinicalbert,tracz-etal-2020-bert,gururangan-etal-2020-dont}. Similarly, for conversational tasks \emph{ConveRT}~\citep{henderson2020} substantially outperforms BERT in neural response selection. Additionally, industrial-scale training on weakly supervised datasets leads to improvements in several NLP tasks~\citep{Snorkel}. 

\section{Method}\label{sec:method}
\subsection{Problem statement}
Given unlabeled instances $\mathcal{D}$, the goal is to automatically cluster utterances into $\mathcal{I}$ classes, which are not known \emph{a priori}. We also assume that we are given labeled instances $\mathcal{D}^{k}$ with $\mathcal{I}_{k}$ known set of intents and $\mathcal{I} \cap \mathcal{I}_{k} \neq \emptyset$. Unlabeled instances may belong to both known intents $\mathcal{I}_{k}$ and unknown ones $\mathcal{I}_{u} = \mathcal{I} \setminus \mathcal{I}_{k}$. 

\subsection{Framework overview}
Our novel intent discovery framework consists of representation learning~\citep{bengio2013representation} and subsequent clustering with K-means~\citep{Lloyd1982LeastSQ}. We propose the following to improve text representations for real-life novel intents discovery in the communication domain:
\begin{itemize}
    \item Efficient initialization with pre-trained encoders, adapted to the e-commerce domain by optimization for weak training signals and conversational structure of the data.
    \item Fine-tuning for the clustering task with state-of-the-art training scheme (i.e., CDAC) adapted to use all the conversational data (i.e., question and answer). \methodname\ is our proposed method to train a conversation structure-aware encoder with three-headed architecture.
\end{itemize}

In the following sections, we describe each component in more detail.

\subsection{Initialization}\label{sec:initialization}
An essential step in the deep learning process is initialization. Proper initialization is crucial in training representations for discovering new intents with clustering. The effectiveness of the existing clustering algorithms depends heavily on the quality of the representation encoder. In this work, we identified this dependency and proposed a generic approach for an efficient encoder pre-training in the conversational domain.

\subsubsection{Domain specific data structure}
We operate in the e-commerce domain with a two-sided marketplace. Customers can seek support by exchanging messages via email or chat. The former are typically longer and include a more formal boilerplate. A dialog may be held between merchants and CX support, buyers and CX support, and directly between buyers and merchants. All messages are written in Polish.

\subsubsection{Domain adaptation}\label{sec:method_domain_adaptation}
We prepared two self-supervised models based on \bertbase~\citep{devlin2019bert} architecture. We started from a general domain encoder \herbert~\citep{mroczkowski-etal-2021-herbert}. We used a training corpus of 68M conversation threads with 184M messages and 8314M words. We included both emails and chats exchanged between all parties (merchants, CX support, and buyers). 

\begin{itemize}
    \item \textbf{\mailbert} is \herbert fine-tuned with Masked Language Model (MLM) objective.
    \item \textbf{\mailbertconvert} is \mailbert\, further fine-tuned on the same dataset but with the mixture of MLM and Conversational Contrastive Loss (CCL)~\citep{henderson-etal-2020-convert}.
\end{itemize}
The details of the training procedure for each of the pre-trained encoders can be found in Appendix~\ref{app:encoders}. 

\subsubsection{Weak supervision}\label{sec:method_weak_supervision}
In the case of email communication exchanged with CX support, every message includes at least one of 512 tags. These labels roughly identify the problem solved. They are assigned by CX consultants often in a noisy manner. We utilized this weak signal and prepared {\bf \mailberttags} encoder in a two-stage process.
Firstly, we finetuned \herbert\ with MLM and Message Threads Structural Objective (MTSO)~\citep{structbert} on all internal communication data (emails and chats). Secondly, we finetuned it on a multi-label classification task on \cxweakly\ dataset that includes 2.5M messages in the email domain exchanged between merchants or buyers and CX support. Details of the training procedure can be found in Appendix~\ref{app:encoders}.

\subsection{\methodname, conversation structure aware encoder}\label{sec:method_conv_cdac}

\begin{figure}[ht]
    \centering
    \includegraphics[width=\columnwidth]{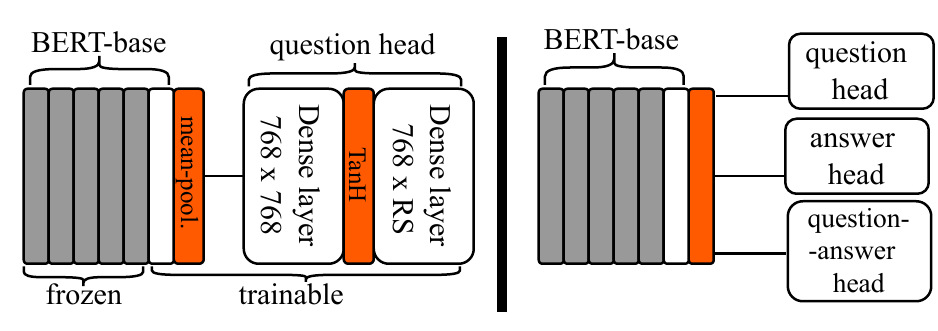}
    \caption{Representation model based on BERT-base encoder used in the discovery pipeline. On the left version with one head. On the right \methodname, our conversational model with three separate trainable heads for the question, answer, and question-answer concatenation. The parameters of the encoder are frozen except for the last transformer block.}
    \label{fig:model}
\end{figure}
As depicted in Fig.~\ref{fig:model}, we used an encoder with \bertbase\ architecture~\cite{devlin2019bert} followed by an average pooling\footnote{Unlike many implementations, the hidden states for padding tokens are not averaged.} and three projection heads with two linear layers and \emph{Tanh} non-linearity in between~\citep{CDAC}.

The three-headed model works with conversational input containing a pair of texts: the user's question and the consultant's answer\footnote{While encoding question and answer, are preceded with special tokens for question and answer.}. Two heads project each input separately, and the third one handles additional signals from the question-answer concatenation into one string of text. Each of the inputs is fed into encoder separately. A common underneath encoder is updated jointly with a gradient from all heads from the total loss given by the weighted average of losses for each head:
\begin{align}\label{eq:lambda}
\mathcal{L}_{\rm \methodname}(X, Y, \theta) = & \lambda_{\rm Q}\cdot \mathcal{L}(X_{\rm Q}, Y, \theta_{\rm Q}) \nonumber\\
&+ \lambda_{\rm A}\cdot \mathcal{L}(X_{\rm A}, Y, \theta_{\rm A}) \nonumber\\
&+  \lambda_{\rm QA}\cdot \mathcal{L}(X_{\rm QA}, Y, \theta_{\rm QA}). \nonumber\\
\end{align}
Here $X=(X_{\rm Q}, X_{\rm A}, X_{\rm QA})$ is the array of inputs  (all examples), i.e. all questions, all answers, all question-answer concatenations respectively. $Y$ are the input labels\footnote{Since we deal with unsupervised/semi-supervised algorithms, some examples are unlabelled.}. $\theta=(\theta_{\rm Q}, \theta_{\rm A}, \theta_{\rm QA})$ is the array of parameter sets for individual inputs \bertbase parameters are shared as depicted in Figure~\ref{fig:model}. The hyperparameters $\lambda=(\lambda_{\rm Q}, \lambda_{\rm A}, \lambda_{\rm QA})$ govern how conversational structure is utilized for any choice of the training scheme, whereas the precise form of the loss terms $\mathcal{L}$ depends on the choice of the training scheme described in Sec.~\ref{sec:method_training_scheme}. For example if we choose $\lambda=(1,0,0)$, and compute $\mathcal{L}$ according to CDAC training scheme, we follow the original CDAC setup with the question field only. By using $\lambda=(0,0,1)$ and computing $\mathcal{L}$ according to CDAC training scheme, we effectively only concatenate question and answer strings and feed it into the model instead of the question string.

In our method \methodname\ for training conversation structure-aware encoder, we trained the representation encoder with uniform heads contribution $\lambda=(\frac{1}{3},\frac{1}{3},\frac{1}{3})$ staring from initializations described in Section~\ref{sec:initialization}. The final representation used for clustering is an embedding from the head for question-answer concatenation.

To speed up training with large batches, we kept the weights of the encoder frozen except for the last transformer layer. The first linear layer keeps the \bertbase\ dimension of the representations (i.e., $768$). The second linear block output dimension is a representation size hyperparameter.

\subsection{Training scheme}\label{sec:method_training_scheme}
Up to this point, we are able to use any framework for finetuning the representation encoder for intent discovery with clustering. With that said, we propose to use two potential approaches for real-world CX communication data.

\paragraph{Static.} In a setup where we do not have any labeled data available, we extract text representation from the pre-trained encoder by average pooling without additional training.

\paragraph{Constrained DAC (CDAC)~\citep{CDAC}.} The method generalizes the Deep Adaptive Clustering (DAC)~\citep{DAC} scheme for partially labeled data and trains with a contrastive loss on both distance-based pseudo-pairs and exact pairs given by intent labels. It is semi-supervised since it utilizes both labeled and unlabeled examples from the train set. We adapted CDAC training scheme to \methodname, our three-headed, conversation structure-aware encoder (see Sec.~\ref{sec:method_conv_cdac}). Details of the DAC method are in Appendix~\ref{appendix:dac}, and details of the CDAC method are in Appendix~\ref{appendix:cdac}.

\section{Evaluation}
We describe our experimental setup for novel intent discovery. We prove the efficiency of the proposed method on real-world communication datasets. To verify gains from different framework components, we present more results in the ablation section (Sec. \ref{sec:ablation}).

\subsection{Real-world internal datasets}\label{sec:datasets}
We used three internal datasets: \purchase, \delivery\, and \retail\ from real traffic to CX support at
\company\ in Polish language. CX consultants manually annotated the datasets with intent labels. Categories of email queries to the CX team are more fine-grained than the widely used  \banking~\citep{casanueva-etal-2020-efficient} dataset. Moreover, such real-world datasets are highly imbalanced, with some intents overlapping. Basic dataset statistics are shown in the Table~\ref{tab:dataset_intro}. The user emails vary in length and style and may contain irrelevant parts. Each dataset includes messages of different quality and specificity ranging from uninformative chit-chat to well-written ones. In datasets, only the first question and direct answer are included, and all further messages from the correspondence thread are omitted. The \purchase\ and \delivery\ cover conversations between buyers and CX consultants. \retail\ is communication between buyers and merchants, so conversation topics and structure are different. We use a stratified 80/10/10 train/val/test split. 

We use two public benchmark English datasets from task-oriented dialog systems: \clinc~\citep{Clinc150} and \banking~\citep{casanueva-etal-2020-efficient} in  Dataset splits follow exactly the experimental setup used in~\citep{zhang_dac2020} in ablation study in Section~\ref{sec:training_scheme} to increase the reproducibilty of our work. In other ablations it is impossible due to missing conversational and weak label signal.

Basic statistics of the datasets are in the Table~\ref{tab:dataset_intro}. Further details are in Appendix~\ref{appendix:datasets}.

\begin{table*}
\centering
    \begin{tabular}{l|cc|cccc|cc}
    \toprule
 & \bf{\#} & \bf{\#}  & \multicolumn{4}{c|}{\bf{\# examples per intent}} &  \multicolumn{2}{c}{\bf{mean length (characters)}} \\
\bf{Dataset} & \bf{intents} & \bf{examples} & \bf{mean} & \bf{min} & \bf{max} & \bf{entropy} & \bf{question} & \bf{answer} \\
\midrule
\tabbanking & 77 & 13.1k & $170{\pm}33$ & 75 & 227 & 0.992 & $60{\pm}40$ & - \\
\tabclinc & 150 & 22.5k & $150{\pm}0$ & 150 & 150 & 0.999 & ${40{\pm}20}$ & - \\
\midrule
 \tabpurchase &  22 & 2.7k &  $121{\pm}50$ & 29 & 240 & 0.972 & $320{\pm}280$ & $1060{\pm}400$ \\
\tabdelivery & 23 & 3.0k & $130{\pm}55$ & 57 & 221 & 0.973& $330{\pm}360$ & $860{\pm}410$ \\
\tabretail & 105 & 13.8k &  $133{\pm}124$ & 22 & 664 & 0.930 & $160{\pm}190$ & $740{\pm}830$ \\
\bottomrule
\end{tabular}
    \caption{Downstream tasks datasets characteristic. Class imbalance is measured by the average number of examples per intent and the normalized Shannon's entropy of the intent distribution (which is 1 for for the perfectly balanced case and lower in case of class imbalance). Further details are in Appendix~\ref{appendix:datasets}}
    \label{tab:dataset_intro}
\end{table*}

\subsection{Experimental setting}\label{sec:experimental_setting}
We build a controlled open-world intent discovery setup, following the setup proposed in~\citep{CDAC,zhang_dac2020}. We prepared novel intents by randomly masking all examples from 50\% of intents in the training set. The remaining intents serve as known intents and are additionally partially masked. We masked 50\% of all remaining examples.
We apply the representation learning framework: we take in-domain encoders described in Section~\ref{sec:method_domain_adaptation} and \ref{sec:method_weak_supervision} and do the fine-tuning step (described in Section~\ref{sec:method_conv_cdac} and \ref{sec:method_training_scheme}). After the training phase, we cluster the whole test dataset with K-means. We performed clustering with the ground truth number of clusters (i.e., the number of intents in the dataset). 

We run experiments with hyperparameters (i.e., representation size, batch size, and learning rate) fixed. We have described the method of their selection in Appendix~\ref{app:fine_tuning}.

We use five random seeds, which govern intent masking and weight initialization. We train the model for 100 epochs on a single machine with NVIDIA V100 GPU. It takes a few hours to run a single fine-tuning experiment for all seeds for a single setting (dataset, training scheme etc.).

\subsection{Metric\footnote{We publish the code for our metrics: \url{https://github.com/allegro/ml/tree/main/publications/intent-discovery-metrics/}}.}
We compute metrics based on cluster ids from K-means algorithm and ground truth labels. 
 The discovery quality is probed with three standard clustering metrics, i.e., Accuracy (ACC) using the Hungarian algorithm, Normalized Mutual Information (NMI), and Adjusted Rand Index (ARI). We also introduce two additional metrics. First, the \emph{binary F1-score} i.e., macro F1-score with a majority vote on cluster label calculated on the whole dataset where all known intents are one class, and all novel intents are the second class. Second, the \emph{macro F1-score} with a majority vote on the cluster label. It turns the clustering quality problem into a multi-label classification.
In the main part of the paper, we report \textbf{AVG} i.e., the average of five metrics over all seeds. AVG increases with clustering quality up to $100\%$. AVG is the primary metric used for model selection. Additionally, to facilitate comparison with other research, the five metrics are listed separately in Appendix~\ref{app:metrics} for all experiments. In Appendix~\ref{app:metrics} we give more details on how we compute metrics or test for statistical significance.
 
\subsection{Results}\label{sec:pretrained_encoders}
Table~\ref{tab:main_results} shows the AVG metric for our best-performing model. Five individual metrics are listed in Table~\ref{tab:main_results_appendix}. We significantly improve intent discovery compared with baselines. \emph{Our} model uses \mailberttags\ (see Section~\ref{sec:method_weak_supervision}) as initialization and is trained with the CDAC scheme. While training, we used both question and answer fields and utilized conversational structure-aware encoder \methodname\ introduced in Sec.~\ref{sec:method_conv_cdac}. The baselines (\emph{Static} and \emph{CDAC}) are based on the general domain \herbert\ encoder and use the question field only. We improved over the second-best CDAC, depending on the dataset, by $8.9$pp to $33$pp. The performance gap of \emph{our} framework to the \emph{CDAC} baseline is greater then the superiority of \emph{CDAC} over the naive baseline, \emph{static} embeddings, which is between 7.7pp and 13.2pp.

\begin{table}[]
    \centering
\begin{tabular}{l|ccc}
\toprule
Method & \tabpurchase &    \tabdelivery &      \tabretail \\
\midrule
Static & $37.0{ \pm }4.1$ &  $31.1{ \pm }1.3$ &  $28.8{ \pm }0.7$ \\
CDAC &   $50.2{ \pm }6.6$ &  $40.9{ \pm }4.5$ &  $36.5{ \pm }1.7$ \\
\midrule
Our &  $\bf{83.2{ \pm }3.2}$ &  $\bf{64.2{ \pm }6.3}$ &  $\bf{45.4{ \pm }4.0}$ \\
\bottomrule
\end{tabular}
    \caption{Static baseline and CDAC representations compared with our framework on novel intent discovery task for real-world data. Our framework combines \mailberttags pre-trained encoder, CDAC training scheme, and \methodname\ method for using the conversation structure. AVG metric averaged over five seeds.}
    \label{tab:main_results}
\end{table}
 
\section{Ablation}\label{sec:ablation}
We attribute the improvement in performance to all three method components: domain adaptation during pre-training with conversational and weak label signal, state-of-the-art training scheme CDAC, and leveraging of conversation structure with our \methodname\ method introduced in Section~\ref{sec:method_conv_cdac}.

\begin{table}[]
    \centering
\addtolength\tabcolsep{-2.5pt}
\begin{tabular}{l|ccc}
\toprule
Initialization &   \tabpurchase &    \tabdelivery &      \tabretail \\
\midrule
\tabherbert & $65.9{ \pm }6.2$ &  $44.7{ \pm }3.7$ &  $37.2{ \pm }2.0$ \\
\tabmailbert &  $66.4{ \pm }6.6$ &  $49.2{ \pm }6.4$ &  $44.2{ \pm }2.2$ \\
\tabmailbertconvert & $73.1{ \pm }8.8$ &  $57.9{ \pm }5.9$ &  $\bf{49.3{ \pm }2.1}$ \\
\tabmailberttags &  $\bf{83.2{ \pm }3.2}$ &  $\bf{64.2{ \pm }6.3}$ &  $45.4{ \pm }4.0$ \\
\bottomrule
\end{tabular}
\addtolength\tabcolsep{2.5pt}
    \caption{Impact of initialization for novel intent discovery task. \methodname\ conversation structure-aware encoder was trained with the CDAC scheme from different initialization. AVG metric averaged over five seeds with standard deviation.}
    \label{tab:initialisation_ablation_results}
\end{table}

\begin{table*}[]
    \centering
\begin{tabular}{l|cc|ccc}
\toprule
Training scheme & \tabbanking & \tabclinc & \tabpurchase & \tabdelivery & \tabretail \\
\midrule
Static & $41.7{ \pm }1.0$ & $55.9{ \pm }1.4$  &  $35.5{ \pm }4.1$ & $31.0{ \pm }2.4$ & $29.6{ \pm }0.8$ \\
DAC & $51.8{ \pm }1.8$ & $64.6{ \pm }1.3$ & $24.1{ \pm }0.7$ & $24.0{ \pm }0.9$ & $27.3{ \pm }4.4$ \\
Supervised & $\bf{65.2{ \pm }2.1}$ & $\bf{73.2{ \pm }0.6}$ & $38.2{ \pm }2.1$ & $33.5{ \pm }2.2$ & $30.1{ \pm }0.5$ \\
CDAC &  $61.8{ \pm }2.8$ & $70.4{ \pm }1.4$ & $\bf{52.9{ \pm }7.3}$ & $\bf{42.3{ \pm }3.6}$ & $\bf{39.2{ \pm }1.2}$\\
\bottomrule
\end{tabular}
    \caption{Evaluation of training schemes for novel intent discovery. We report AVG metric averaged over five seed with standard deviation. Models use \bertbase\ (English datasets) or \mailbert\ (Polish datasets) encoder and question input only. The best results are in bold.}
    \label{tab:clustering_results}
\end{table*}

\subsection{Initialization}\label{ablation_init}
In this section, we show the effect of initialization on the novel intent discovery task. We trained a conversation structure-aware encoder with a CDAC scheme using four different initializations. 

AVG metric is reported in Table~\ref{tab:initialisation_ablation_results} and individual metrics are shown Table~\ref{tab:initialisation_ablation_results_appendix}. Comparing \mailbert\ with \herbert, we can see that domain-adapted initialization improves $1$ to $7$pp for discovering new intents. Further adaptation of the starting encoder with the loss of ConveRT improves at least $5$pp. Summarizing \mailbert\ and \mailbertconvert\ initializations bring gains for all internal datasets.
For the CX domain (\purchase\ or \delivery ), the best initialization was provided by \mailberttags. Pre-training with weak labels introduced additional training information that turned out to be transferable for the downstream task. The simultaneous drop in quality on the \retail\ dataset originating from the domain for which we did not have noisy labels confirms this phenomenon. 
 
\subsection{Training schemes}\label{sec:training_scheme}
We compare two training schemes \emph{Static}, and \emph{CDAC} from Sec.~\ref{sec:method_training_scheme} with two additional baseline methods \emph{DAC} and \emph{Supervised}. For \emph{Supervised} training scheme, we use Large Margin Cosine Loss (LMCL)~\citep{8578650} to learn representation from labels. We discard unlabeled data from the train set. We train the models for all four schemes with question input only and \bertbase~\cite{devlin2019bert} for English and \mailbert\ for Polish datasets.

This ablation study is the only case when we can use two public benchmark English datasets from task-oriented dialog systems: \clinc~\citep{Clinc150} and \banking~\citep{casanueva-etal-2020-efficient}. Unfortunately, public benchmark datasets lack the answer data, a large amount of unlabeled data, and weak labels. However, including them in this ablation study increases the reproducibility of our work and brings interesting insights.

AVG metric is reported in Table~\ref{tab:clustering_results} and individual metrics can be found in Table~\ref{tab:clustering_results_appendix}. For all datasets, there is a gain from using intent labels (\emph{Supervised} and \emph{CDAC}). For public datasets among unsupervised methods, DAC outperforms static representations. However, supervised training is better than semi-supervised CDAC. The results are the opposite for the internal datasets. DAC is better than static representations, and semi-supervised CDAC is better than supervised training. We hypothesize that different real-world and benchmark datasets results might be due to dataset quality and size differences. In general, benchmark datasets are larger and more balanced. Moreover, mail messages from real-world e-commerce are longer and noisier on average. It is an open question how this trend holds for other real-life datasets. 

To sum up, there is a gain from intent labels for all datasets. Optimal solutions for public benchmarks and real-world internal datasets differ. CDAC is the best training scheme that uses intent labels for internal datasets.

\subsection{Conversational structure}\label{ablation_conv}
We examine if any further gains in performance can be obtained from incorporating the answer field signal. 
We conduct experiments only on the internal datasets. We use only the best training scheme, i.e., \emph{CDAC}. We examine four training configurations: only question representation \emph{Q} trained with $\lambda=(1,0,0)$, only answer representation \emph{A} trained with $\lambda=(0,1,0)$, question-answer concatenation \emph{QA concatenation} trained with $\lambda_{3}=(0,0,1)$, using question and answer in a simpler two-headed model \emph{QA two heads} trained with $\lambda=(\frac{1}{2},\frac{1}{2},0)$ and full three-headed conversational model \methodname\ trained with $\lambda=(\frac{1}{3}, \frac{1}{3},\frac{1}{3})$ described in detail in section Sec.~\ref{sec:method_conv_cdac}.\footnote{For multi-headed encoders, we chose the best of all possible final representations (output from any head, or concatenations of outputs from multiple heads).}

AVG metric is reported in Table~\ref{tab:clustering_results_qa} and individual metrics can be found in  Table~\ref{tab:clustering_results_qa_appendix}. The answer alone performs worse than the question alone. We hypothesize that it is due to many non-informative generic answers\footnote{e.g., \emph{Thank you for your message. Let me check some details and reply later.}}. Perhaps for other real-world datasets consultant's answer may be superior to the user's questions. Passing only the question signal is a strong baseline. Let us check if it is possible to incorporate signals from both question and answer fields in a way that improves performance over \emph{Q}, question field only baseline. The most straightforward extension, \emph{QA concatenation}, which requires only inputting different inputs to the same model is slightly better but does not pass the statistical significance test. The same goes for the more sophisticated \emph{QA two heads} variant. Only our method \methodname, a three-headed encoder is better than \emph{Q} with statistical significance.  Incorporating both question and answer signal leads to further improvements.

To sum up, after examining multiple ways to include the conversational signal, we conclude that our method \methodname\ with a three-headed encoder improves the performance by $5$ to $13.5$pp.
\begin{table}[]
    \centering
\addtolength\tabcolsep{-3.0pt}
\begin{tabular}{l|ccc}
\toprule
 & \tabpurchase &    \tabdelivery &      \tabretail \\
\midrule
\emph{Q} & $52.9{ \pm }7.3$ &  $42.3{ \pm }3.6$ &  $39.2{ \pm }1.2$ \\
\emph{A} & $51.7{ \pm }5.5$ &  $37.6{ \pm }4.5$ &  $30.5{ \pm }1.5$ \\
\hline
\emph{QA concat.} & $55.1{ \pm }3.8$ &  $47.3{ \pm }3.4$ &  $43.4{ \pm }3.1$ \\
\emph{QA two head.} & $56.4{\pm}5.9$ & $46.9{\pm}5.4$ & $40.2{\pm}1.7$ \\
\methodname & $\bf{66.4{ \pm }6.6}$ &  $\bf{49.2{ \pm }6.4}$ &  $\bf{44.2{ \pm }2.2}$ \\
\bottomrule
\end{tabular}
\addtolength\tabcolsep{3.0pt}
    \caption{Evaluation of conversational structure for novel intent discovery. We report AVG metric averaged over five seed runs with standard deviation. Models use \mailbert\ initialization, CDAC training scheme, and various inputs, i.e., question \emph{Q}, answer \emph{A}, or both fields (QA) in three model variants; \emph{QA concatenation}, \emph{QA two heads}, and \methodname. The best results are in bold.}
    \label{tab:clustering_results_qa}
\end{table}

\section{Commercial deployment}\label{sec:deployment}
\subsection{Production pipeline overview}
\begin{figure}[ht]
    \centering
    \includegraphics[width=\columnwidth]{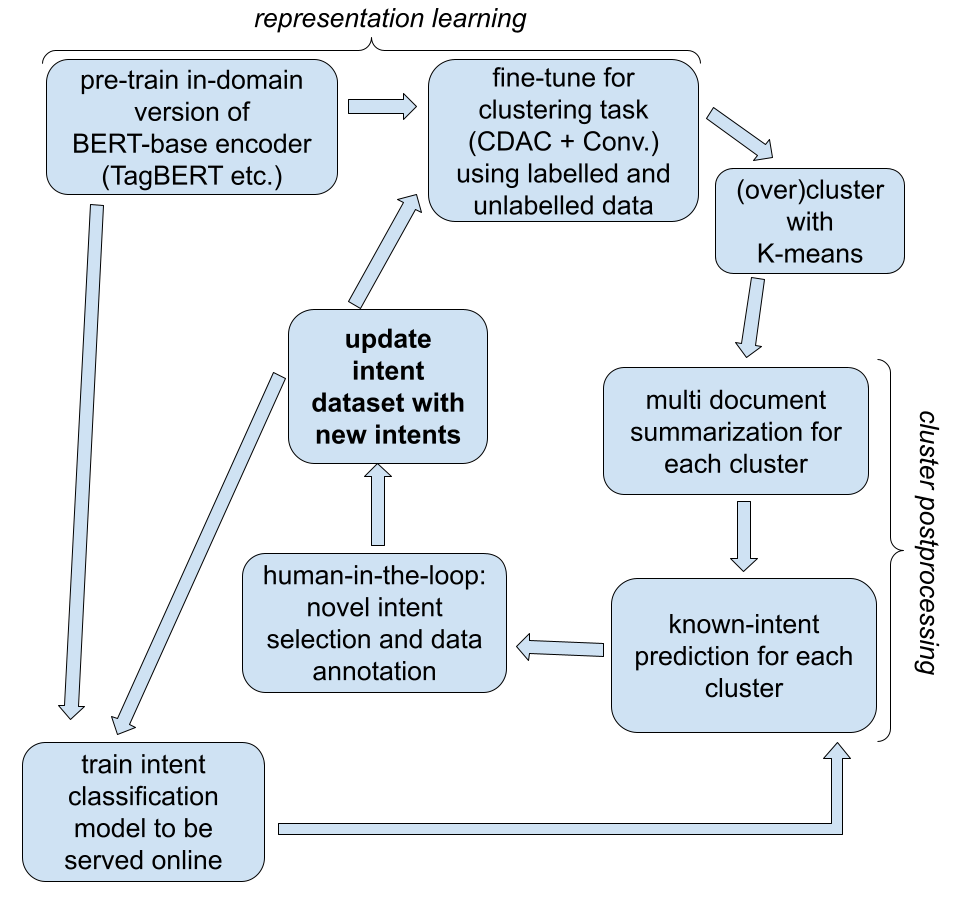}
    \caption{Intent discovery pipeline deployed at \company\ with human-in-the-loop carrying out the novel intent selection and data annotation. Representation learning components are subject to experiments in this paper. The main outcome of the pipeline is an updated intent detection dataset, which can be used to train a better intent classification model.}
    \label{fig:intent_pipeline}
\end{figure}
The method we described and verified experimentally is a part of a larger multi-component system for continuous intent discovery deployed commercially, shown in Fig.~\ref{fig:intent_pipeline}. Here we briefly list the major components of our production pipeline to give the bigger picture:

\begin{enumerate}
    \item \emph{Representation learning.} Representation learning plays a core role in our pipeline. This component is subject to experiments in this paper and consists of two subcomponents:
\begin{enumerate}
\item \emph{In-domain pre-training of encoders.} Encoders with \bertbase\ architecture are pre-trained on large chunks of historical data. We include additional signals such as conversational structure (i.e. question and answer) and weak label signal (Section~\ref{sec:method_domain_adaptation} and \ref{sec:method_weak_supervision}). The encoders are reused for the intent classification model.
\item \emph{Fine-tuning for the clustering task.} We further train in-domain encoders. If there exists annotated data, we use semi-supervised CDAC with \methodname\ (Section~\ref{sec:method_conv_cdac}). Otherwise, we use static embeddings.
\end{enumerate}
\item \emph{(Over)clustering with K-Means.} We cluster representations to discover intent groups in the data. The number of novel intents is required by K-Means. We overestimate this value as it is less time-consuming to manually merge clusters with the same intent.
\item \emph{Cluster postprocessing.} Various postprocessing steps make  analyzing the clusters by the human annotators more efficient:
\begin{enumerate}
\item \emph{Multi-document summarization.} The summarization module, provides human-readable candidates for the intent name instead of cluster ids. First, we train a logistic regression classifier with bag-of-words features to predict cluster ids. Then, we identify the most informative sentence in each message using the classifier coefficients~\citep{angelidis-lapata-2018-summarizing}. Finally, we select the five most central sentences across all messages \citep{zheng-lapata-2019-sentence}.
 \item \emph{Known intent prediction.} We need to distinguish clusters with known intents from clusters with potentially novel intents. Since the labeled messages are typically a small subset of the training dataset, we infill intents for the unlabeled examples with an intent classifier and present this information to human annotators.
 \end{enumerate}
\item \emph{Novel intent selection and data annotation.} Human annotators manually analyze all discovered clusters and choose which novel intents to include in the taxonomy. They annotate all messages from clusters to be included in the labeled dataset to ensure the high coherence of newly discovered intents.
\end{enumerate}
CX intent dataset updated with new intent is the end product of our intent discovery pipeline. Its primary purpose is to train an intent classifier to be served in real-time to CX consultants. It is a complex pipeline of its own. It has similar architecture to the representation learning model in the intent discovery pipeline and it reuses pre-trained encoders. Even though the consultant's answer and the consultant's weak label are not known at the serving time of the intent classification model, we leverage these signals to build a better intent dataset and directly train a better intent classification model.
\subsection{Commercial benefits case study}
Thanks to the deployed pipeline, we doubled the number of defined intents for customer support within one year. Initially, the taxonomy consisted of 100 classes manually defined by the CX consultants. The commercial deployment of the intent discovery pipeline happened at the moment when the domain experts failed to find any new intents manually. Roughly 50 new intents were discovered thanks to our intent discovery pipeline. The selected clusters were reasonably pure: over 90\% (mean and median) of examples from the selected clusters were labeled as the given intent. Additional examples for the new intents were further added (active learning etc.) and at the moment, the examples from the clustering process are at least 40\% of all examples for 50 automatically discovered intents. Currently, after extending our taxonomy from other sources as well, our taxonomy has roughly 180 intents.

In addition, the pipeline decreased the time required to define novel intents from weeks to days with the additional benefit of analyzing several-fold more messages. The more comprehensive taxonomy significantly impacts the total benefit from the automation process, improves user experience by providing faster responses, and saves the cost of hiring additional CX consultants.

\section{Conclusions}
This paper describes an intent discovery pipeline deployed on a large e-commerce platform. The access to real-life datasets allows extending the established intent discovery models to better leverage vast amounts of unlabelled data, its conversational structure, and additional signals like weak labels. In particular, we learn the following lessons:
\begin{enumerate}
    \item Among multiple ways to handle conversational data, \methodname, our generalization of the CDAC model to a three-headed encoder to use all available conversational data (i.e., question and answer) increases the performance of the intent discovery pipeline the most. See Section~\ref{ablation_conv}.
    \item The significant gains also come from pre-training the encoder on an unlabelled in-domain dataset with conversational structure and weak labels (\mailberttags). See Section~\ref{ablation_init}. Therefore, we recommend a system architecture that enables weak labeling by the consultants by design.
    \item Even though the consultant's answer and weak labels are not available at the serving time of the intent classification model, they can be used offline for novel intent discovery to build a better dataset and directly improve the intent classification. It happened for our comercially deployed pipeline. See Section~\ref{sec:deployment}.
    \item Gains from incorporating additional signals (\methodname\ method, \mailberttags) are larger than gains from using state-of-the-art methods (CDAC) on datasets without additional signals. See Section~\ref{sec:pretrained_encoders}. We advocate for a shift both in construction and research on intent detection datasets. 
\end{enumerate}
\section{Limitations}
We are aware of two major factors that may affect the generality of our research: shortcomings of the simulated novel intent discovery setup and the assumption that intent detection is a classification problem.
\paragraph{Simulated experiments.} In the experimental section, we use small, entirely annotated datasets to analyze different design choices of the representation learning component. We naturally include only already discovered intents (does not mean these are all possible). Our masking procedure that follows research papers~\citep{CDAC,zhang_dac2020} has three drawbacks. Firstly, when we mask most of the dataset, we effectively do few-shot learning, whereas, in reality, the amount of annotated data is much larger. The observed differences between design choices may be mitigated once more data is available. Secondly, real class imbalance may not be reflected in the experimental dataset due to the annotation procedure. Lastly, the ratio between batch size and dataset size is much smaller for real datasets since, in general, we are training with a large amount of unannotated data. It directly affects batch-based pair statistics when using a random sampler in CDAC algorithm. The chance that annotated examples will be present in the batch is low, and effectively we are almost entirely learning from pseudo-pairs during the semi-supervised stage.  
\paragraph{Intent detection as classification.} We treat the intent discovery as classification i.e. each utterance has only one intent. In reality, users may have more than one goal that transforms the problem into a multi-label scenario. Naturally, we could treat multi-label examples as yet another class, but we do not explore their influence on pipeline performance since they were in a significant minority.
\section*{Acknowledgements}
We thank Bartosz Ludwiczuk, Tomi Wójtowicz, Karol Grzegorczyk
for valuable discussions and contributions to the source code.

\bibliography{anthology,custom}
\bibliographystyle{acl_natbib}

\appendix
\include{appendix}

\end{document}

%% file: appendix.tex
\section{Dataset details}\label{appendix:datasets}
We further describe real-world internal datasets introduced in \ref{sec:datasets} and compare them to public benchmark datasets. Table~\ref{tab:intent_sample} exemplifies the domain diversity of the datasets: it contains three sample intent names per dataset.

\begin{figure*}[ht]
    \centering
    \includegraphics[width=\textwidth]{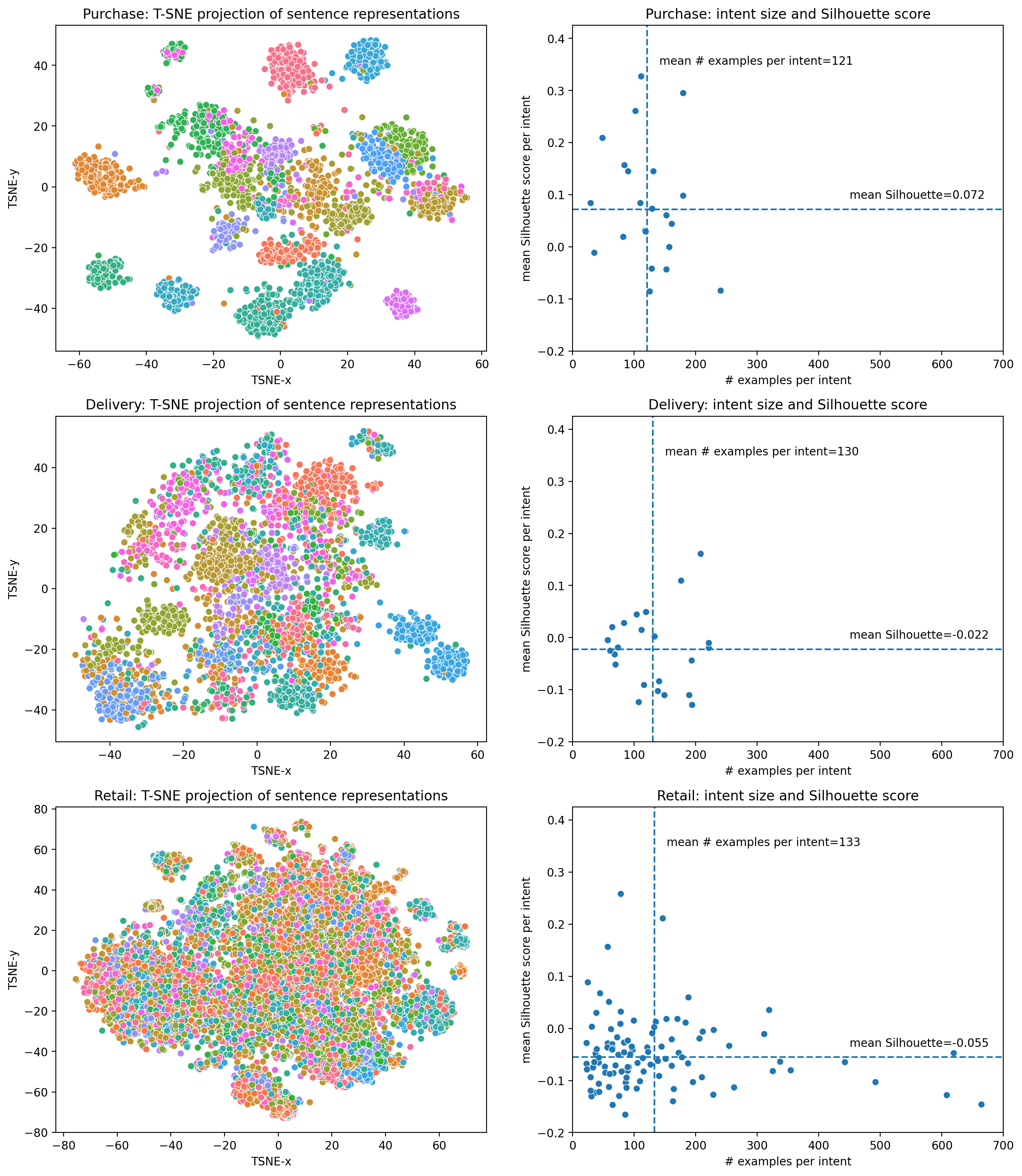}
    \caption{Internal dataset visualization. On the left we visualize t-SNE mapping of sentence representations to 2 dimensions. Different colors indicate different intent labels, each point corresponds to a single example in the dataset. On the right there is a scatter plot of intent sizes and Silhouette score per intent. Each point corresponds to one intent in the dataset. Silhouette score values are in the range from -1 to 1. 1 indicates perfect clustering, and 0 indicates overlapping clusters. The visualizations show the initial difficulty of the clustering task on general domain pre-trained models.}
    \label{fig:vis_internal}
\end{figure*}

\begin{figure*}[ht]
    \centering
    \includegraphics[width=\textwidth]{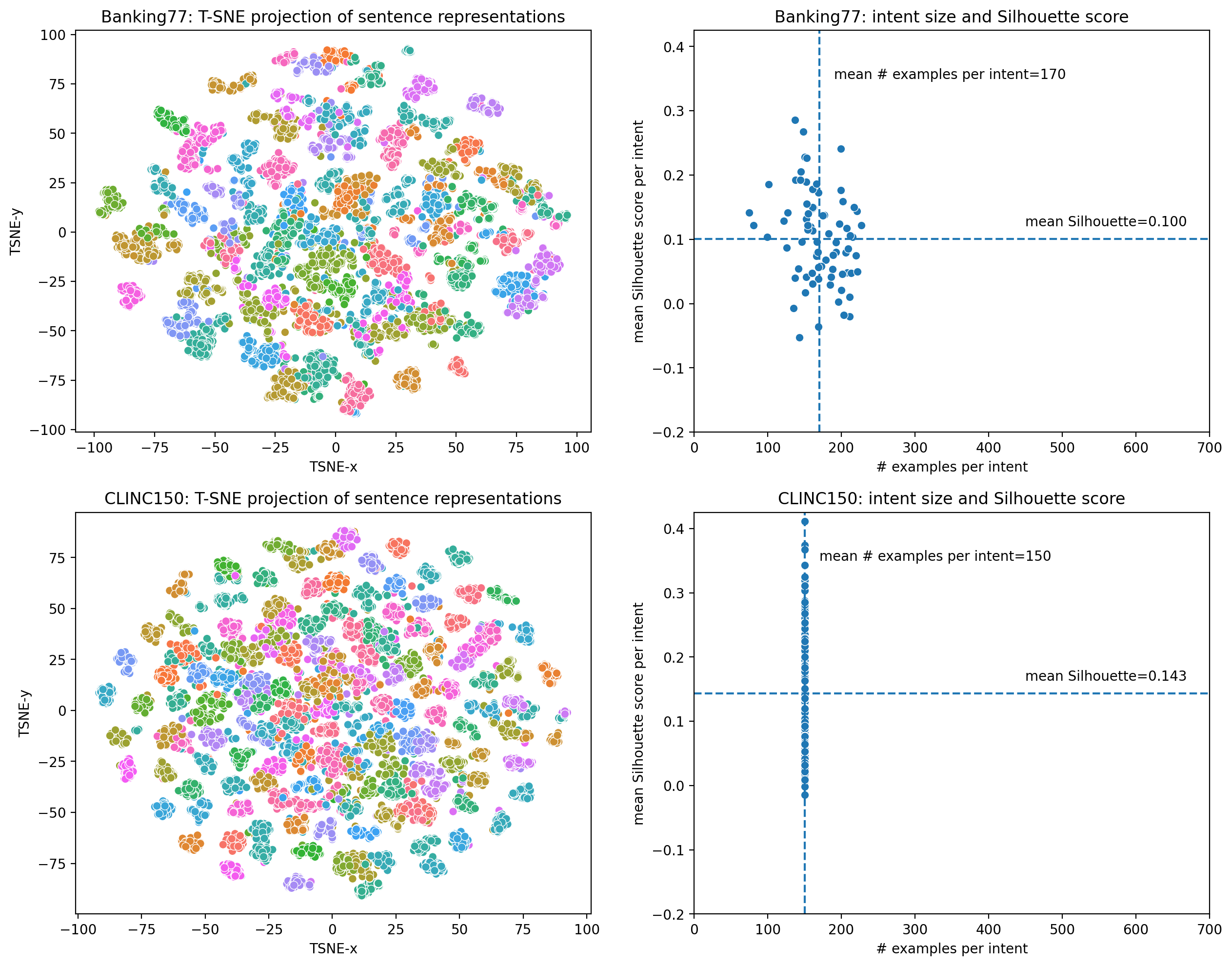}
    \caption{Public dataset visualization. On the left we visualize t-SNE mapping of sentence representations to 2 dimensions. Different colors indicate different intent labels, each point corresponds to a single example in the dataset. On the right there is a scatter plot of intent sizes and Silhouette score per intent. Each point corresponds to one intent in the dataset. Silhouette score values are in the range from -1 to 1. 1 indicates perfect clustering, and 0 indicates overlapping clusters. The visualizations show the initial difficulty of the clustering task on general domain pre-trained models.}
    \label{fig:vis_public}
\end{figure*}

We visualize the datasets. We use publicly available pre-trained models to enable simple visual comparisons between our real-world internal datasets and any other datasets. 
Sentence-BERT produces English sentence embeddings by fine-tuning on semantic textual similarity STS pairs~\citep{reimers-gurevych-2019-sentence}. We use a variation of Sentence-BERT trained from MPNet~\citep{NEURIPS2020_c3a690be}. Polish version has been obtained following knowledge distillation procedure~\citep{reimers-gurevych-2020-making,polish-nlp-resources}. 
\footnote{Package \texttt{sentence-transformers}, available at \url{https://sbert.net}, is used with models \texttt{all-mpnet-base-v2} or \texttt{sdadas/st-polish-paraphrase-from-mpnet} for English and Polish respectively.} We compute sentence embeddings for the question field or if the answer field is present, for question-answer concatenation. For each example, we compute a partial Silhouette score (using ground truth intents as cluster labels) and average it per intent. Silhouette score, designed originally for evaluating the clustering quality, takes into account the mean intra-cluster distance and the mean nearest-cluster distance for each example. We plot 2D t-SNE mappings of the embeddings, Silhouette score per intent~\footnote{\url{https://scikit-learn.org/}}, and intent sizes in Figures~\ref{fig:vis_internal} and~\ref{fig:vis_public} to visualize the datasets and the initial difficulty of the clustering task on general domain pre-trained models.
\begin{table}[]
    \centering
    \begin{tabular}{l|c}\hline
        \bf{Dataset} & \bf{Three sample intent labels} \\
        \hline
        \multirow{3}{*}{\tabbanking} & 1. Cash withdrawal charge \\
        & 2. Getting spare card \\
        & 3. Request refund \\
        \hline
        \multirow{3}{*}{\tabclinc} & 1. Transactions \\
        & 2. Next song \\
        & 3. International fees \\
        \hline
        \multirow{3}{*}{\tabpurchase} & 1. I have a technical problem. \\
        & \linebreakcell{2. When will my Smart!\\ be active?} \\
        & \linebreakcell{3. How to withdraw from\\ the auction?} \\
        \hline
        \multirow{3}{*}{\tabdelivery} & \linebreakcell{1. I didn't pick up my parcel\\and I'm asking for a refund.} \\
        & \linebreakcell{2. How to withdraw\\from the contract?} \\
        & \linebreakcell{3. I want to use Buyers\\Protection Program.} \\
        \hline
        \multirow{3}{*}{\tabretail} & \linebreakcell{1. When will the sale of\\the offer start?} \\
        & \linebreakcell{2. I have a problem with the cust-\\omer service for my purchase.} \\
        & 3. Is the product prepackaged? \\
        \hline
    \end{tabular}
    \caption{Domain diversity of labeled datasets used for novel intent discovery experiments. Three sample intent names per datasetare given.}
    \label{tab:intent_sample}
\end{table}

\begin{table}[]
    \centering
\addtolength\tabcolsep{-1.5pt}
\begin{tabular}{l|cc|ccc}
\toprule
\bf{Dataset} &  \rot{\tabbanking} &  \rot{\tabclinc} &  \rot{\tabpurchase} &  \rot{\tabdelivery} &  \rot{\tabretail} \\
\midrule
\bf{Representation size} &          256 &        256 &            32 &            32 &          64 \\
\bf{Batch size}          &          128 &        128 &            16 &            32 &          16 \\
\bf{\# intents} & 77 & 150 & 22 & 23 & 105 \\
\bottomrule
\end{tabular}
    \caption{Optimal representation size and batch size vs. a number of annotated intents in the datasets.}
    \label{tab:representation_batch}
\addtolength\tabcolsep{1.5pt}
\end{table}

\section{Training schemes}
\subsection{Deep Adaptive Clustering (DAC)}\label{appendix:dac}
It was introduced in~\citep{DAC} for the Computer Vision domain but is easily extended to text. Originally, output representation was interpreted as a probability distribution over unique classes, i.e., they used $L_{2}$ normalized features with positive elements. We relaxed this condition and trained real-valued representation for any clustering algorithm. The representation size doesn't have to match a unique number of classes in the dataset (unknown in real scenarios). For a pair of examples $i$, $j$ the loss function $\mathcal{L}_{ij}$ is
 \begin{equation}
     \mathcal{L}_{ij} = -R_{ij}\log S_{ij} - (1 - R_{ij}) \log (1 - S_{ij}),
 \end{equation}
where ($R_{ij} = 1$) for positive pairs and ($R_{ij} = 0$) for negative pairs and $S_{ij}$ is cosine similarity of representations. The pseudo-label matrix $R$ is defined in an online fashion for every pair of examples in a batch using current model predictions i.e.
\begin{equation}\label{eq:pseudo_label_matrix}
    R_{ij} =
    \begin{cases}
    1, & \text{if } S_{ij} \geq u(\lambda), \\
    0, & \text{if } S_{ij} < l(\lambda), \\
    \text{None}, & \text{otherwise},
    \end{cases}
\end{equation}
where $u(\lambda)$ and $l(\lambda)$ are upper and lower thresholds. Pairs between the thresholds do not take part in the training. This is compensated by adding penalty term $u(\lambda) - l(\lambda)$ to the final loss. The thresholds are updated every epoch according to the formula
\begin{align*}
    u(\lambda) & = 0.95 - \lambda, \\
    l(\lambda) & = 0.455 + 0.1 \cdot \lambda,
\end{align*}
where update rule for $\lambda$ every epoch is $\lambda = \lambda + 1.1 \cdot 0.009$~\citep{DAC}. We start with $\lambda = 0$. The training ends when $u(\lambda) = l(\lambda)$. The training resembles curriculum learning: we start with confident examples with very large or low cosine similarity and then introduce more uncertainty. The penalty term also reflects our confidence since it controls the strength of gradient updates.

\subsection{Constrained DAC (CDAC)}\label{appendix:cdac}
This extension of DAC to a semi-supervised scenario was introduced in~\citep{CDAC}. In unsupervised case, we only use contrastive objective with pseudo-labels. Once we have annotated examples, we define true positive and negative pairs with labels. The label matrix $R$ has now pseudo-label part~\eqref{eq:pseudo_label_matrix} and exact part
\begin{equation}\label{eq:exact_label_matrix}
    R_{ij} = 
    \begin{cases}
    1, & \text{if } y_i = y_j, \\
    0, & \text{if } y_i \neq y_j,
    \end{cases}
\end{equation}
where $y_{i}$ denotes encoded label for $i$-th example. Since our batch now includes annotated and unannotated examples, we need to redefine pseudo-labels. We consider three cases. Firstly, pseudo-labels can be defined only among unannotated examples. Secondly, we can allow pseudo-labels between pairs of annotated and unannotated examples. Lastly, we can define pseudo-labels for all possible pairs, including a scenario where pseudo-labels are defined among annotated pairs. We chose the second scenario.

Additional modification is alternating training. Even epochs use only annotated data and no threshold penalty. Odd epochs use the whole dataset and pseudo-label matrix as well as exact. The loss in the supervised phase is additionally scaled by the $\delta \geq 1$ hyperparameter to control the weight put on annotated data.

\section{Metrics\footnote{We publish the code for our metrics: \url{https://github.com/allegro/ml/tree/main/publications/intent-discovery-metrics/}}.}\label{app:metrics}
We choose metrics for our experiments. Three clustering metrics measure the separation of novel intents from each other:
\begin{itemize}
    \item \textbf{Accuracy (ACC)} measures clusters purity. Cluster and ground-truth labels are matched with the Hungarian algorithm.
    \item \textbf{Normalized Mutual Information (NMI)} specifies the amount of uncertainty about class labels given cluster labels.
    \item \textbf{Adjusted Rand Index (ARI)} checks for all sample pairs whether their assigned and ground truth labels are the same.
\end{itemize}
ACC, NMI, and ARI  are calculated only on examples with a novel intent as a ground truth label.

The separation of the novel from the known intents is measured by:
\begin{itemize}
    \item \textbf{Binary F1-score}. It is a macro F1-score with a majority vote on the cluster label calculated on the whole dataset where all known intents are one class and all novel intents are the second class.
\end{itemize}
Last but not least, there is a metric that measures both the separation between novel intents and the separation of the novel from the known:
\begin{itemize}
    \item \textbf{Macro F1-score} with majority vote on cluster label. It turns the clustering quality problem into multi-label classification.
\end{itemize}
The macro average is calculated only for novel intents. Examples with any ground truth label may be included\footnote{See: \url{https://scikit-learn.org/stable/modules/generated/sklearn.metrics.f1_score.html}}.

All metrics increase with clustering quality up to $100\%$. We use five random seeds, which govern intent masking and weight initialization. In the main part of the paper, we report \textbf{AVG} i.e., the average of five metrics listed above (which are correlated variables) overall seeds. AVG is the primary metric used for model selection. Whenever in doubt, we confirm that the difference between AVG metrics is statistically significant with correlated T-Test with a p-value=5\% threshold. Additionally, to facilitate comparison with other research, for all experiments, the five metrics are listed separately in Appendix. 

\section{Initial fine-tuning}\label{app:fine_tuning} 
We start our experiments with fine-tuning representation size, batch size, and learning rate hyperparameters for the CDAC training scheme\footnote{We focus on CDAC encouraged by initial good results for CDAC and high cost of fine-tuning each training scheme separately.}. For every dataset, we optimize the hyperparameters in two steps: selecting optimal representation size via grid search over the representation sizes \{16, 32, 64, 128, 256\} and learning rates \{1e-05, 5e-05, 1e-04\} and then selecting the optimal learning rate and batch size via grid search over batch sizes \{16, 32, 64, 128, 256, 512\} and the same learning rates as step 1. Tab.~\ref{tab:representation_batch} shows the relation of the selected hyperparameters to the number of intents. The selected hyperparameters are later fixed in the experiments. Additionally, to improve training stability, we perform an additional learning rate search again within values \{1e-05, 5e-05, 1e-04\} for every setup which uses \methodname\ method separately.

\section{Pre-trained encoders (details)}\label{app:encoders}
To leverage large amounts of historical data, we compare four self-supervised encoders, and one supervised trained on conversational data. The training procedure for each encoder is described in detail below for reproducibility. The encoders are used for experiments in Sec.~\ref{sec:pretrained_encoders}.

\paragraph{\herbert} State-of-the-art \bertbase\ language model for Polish~\citep{mroczkowski-etal-2021-herbert} trained with Masked Language Model (MLM) objective.

\paragraph{\mailbert} The model is a result of further fine-tuning \herbert\ on internal unsupervised conversational data. The single training example contains a conversation thread clipped to 512 tokens. We always clip threads to a random subsequence of whole consecutive utterances to persist in a conversational context. \mailbert\ is trained with the MLM objective for 100k steps with the linearly decaying learning rate schedule (peak value  1e-05) and the batch size of $224$. The training on four NVIDIA A100 GPUs lasted 2 days. 

\paragraph{\mailbertconvert} The model is a result of further fine-tuning of the \mailbert\ on the same data but with the mixture of two objectives, MLM loss with the ratio of $0.2$ and Conversational Contrastive Loss (CCL). Following ConveRT~\citep{henderson-etal-2020-convert} we leverage the structure of the conversations with alternately exchanged utterances in a metric learning setup. Positive examples are consecutive messages from a single conversation, and negatives come from answers within the training batch. To reduce the overfitting to specific utterances, we use label smoothing with the value of $0.2$ (same as~\citep{henderson-etal-2020-convert}). To utilize conversational data structure, we add two projection heads on top of the \mailbert\ encoder, one for the question and answer representations\footnote{Answers in our data come from two sources: CX consultants and sellers.}. \mailbertconvert\ is trained for the 280k steps with the peak learning rate 1e-05 and the batch size of $448$. The training on four NVIDIA A100 GPUs lasted 4 days.
\paragraph{\mailberttags} The model is trained in two-stage fine-tuning of the first version of \herbert~\citep{rybak-etal-2020-klej}. In the first stage, we fine-tune the model on internal unsupervised conversational data. We use MLM objective and Message Threads Structural Objective (MTSO). MTSO is Sentence Structural Objective~\citep{structbert} tailored to the conversation domain. During training, we swap messages with respect to threads instead of swapping sentences with respect to documents. \mailberttags\ is trained for 100k steps with a batch size of $640$ and a peak learning rate 8e-05.

In the second stage, we fine-tune the model on the multi-label classification task. The model predicts several of the 512 classes for each thread. The noisy and highly imbalanced labels come from tags that CX consultants add to the conversation threads, roughly identifying the problem solved. The training dataset contains 2.5M messages. \mailberttags\ is trained for 38k steps with a peak learning rate of 1.6e-04 and a batch size of $512$. The training on sixteen NVIDIA P100 GPUs lasted 8 hours.

\section{Results (details)}\label{app:metrics}

\begin{table*}[]
    \centering
\begin{tabular}{l|ccccc|ccccc|ccccc}
\hline
\bf{Dataset} & \multicolumn{5}{c|}{\tabpurchase} & \multicolumn{5}{c|}{\tabdelivery} & \multicolumn{5}{c}{\tabretail} \\
\hline
Method  & \rot{macro F1} &  \rot{ACC} &  \rot{NMI} & \rot{ARI} & \rot{binary F1} & \rot{macro F1} & \rot{ACC} & \rot{NMI} & \rot{ARI} & \rot{binary F1} & \rot{macro F1} & \rot{ACC} & \rot{NMI} & \rot{ARI} & \rot{binary F1}  \\
\hline
Static & 23 & 39 & 45 & 17 & 62 & 19 & 28 & 37 & 8 & 64 & 10 & 20 & 47 & 5 & 62 \\
CDAC & 33 & 50 & 61 & 30 & 77 & 27 & 39 & 50 & 16 & 72 & 15 & 32 & 57 & 10 & 67 \\
\hline
Our & \bf{75} & \bf{83} & \bf{88} & \bf{78} & \bf{92} & \bf{49} & \bf{64} & \bf{72} & \bf{56} & \bf{81} & \bf{19} & \bf{42} & \bf{65} & \bf{31} & \bf{70} \\
\hline
\end{tabular}
    \caption{Static baseline and CDAC representations compared with our framework on novel intent discovery task for real-world data. Our framework combines \mailberttags pre-trained encoder, CDAC training scheme, and \methodname\ method for using the conversation structure. Individual metrics averaged over five seeds.}
    \label{tab:main_results_appendix}
\end{table*}

\begin{table*}[]
    \centering
\begin{tabular}{l|ccccc|ccccc|ccccc}
\hline
\textbf{Dataset} & \multicolumn{5}{c|}{\tabpurchase} & \multicolumn{5}{c|}{\tabdelivery} & \multicolumn{5}{c}{\tabretail} \\
\hline
Initialization & \rot{macro F1} &  \rot{ACC} &  \rot{NMI} &  \rot{ARI} & \rot{binary F1} & \rot{macro F1} & \rot{ACC} &  \rot{NMI} &  \rot{ARI} & \rot{binary F1} & \rot{macro F1} &  \rot{ACC} &  \rot{NMI} &  \rot{ARI} & \rot{binary F1}  \\
\hline
\tabherbert & 53 & 66 & 73 & 53 & 84 & 27 & 44 & 49 & 25 & 78 & 18 & 33 & 57 & 10 & 68 \\
\tabmailbert & 54 & 67 & 74 & 52 & 86 & 33 & 46 & 58 & 32 & 77 & 17 & 42 & 65 & 27 & 70 \\
\tabmailbertconvert & 60 & 74 & 83 & 67 & 83 & 46 & 57 & 64 & 41 & \bf{81} & \bf{20} & \bf{48} & \bf{71} & \bf{36} & \bf{72} \\
\tabmailberttags & \bf{75} & \bf{83} & \bf{88} & \bf{78} & \bf{92} & \bf{49} & \bf{64} & \bf{72} & \bf{56} & \bf{81} & 19 & 42 & 65 & 31 & 70 \\
\hline
\end{tabular}
    \caption{Impact of initialization for novel intent discovery. \methodname\ conversation structure-aware encoder was trained with the CDAC scheme from different initialization. Individual metrics averaged over five seeds.}
    \label{tab:initialisation_ablation_results_appendix}
\end{table*}

\begin{table}[]
    \centering
\begin{tabular}{ll|cc|ccc}
\hline
& \bf{Dataset} &  \rot{\tabbanking} &  \rot{\tabclinc} &  \rot{\tabpurchase} &  \rot{\tabdelivery} &  \rot{\tabretail} \\
\hline
\multirow{5}{*}{\rot{Static}} & macro F1 &     30 &         44 &            21 &            17 &          11 \\
& ACC & 33 & 49 & 36 &            30 &          22 \\
& NMI & 55 &         75 &          41 &            36 &          48 \\
& ARI & 23 & 36 &          14 &             9 &           6 \\
& binary F1 &         68 &       75 &            65 &            63 &          62 \\
\hline
\multirow{5}{*}{\rot{DAC}} & macro F1 & 42 &         55 &            13 &            12 &          10 \\
& ACC & 45 & 58 &  22 &            20 &          17 \\
& NMI & 64 & 81 &          30 &            28 &          46 \\
& ARI & 35 & 49 &           0 &             1 &           3 \\
& binary F1 &  73 &       80 &            55 &            59 &          61 \\
\hline
\multirow{5}{*}{\rot{Supervised}} & macro F1  &  55 &         64 &            22 &            19 &          11 \\
& ACC & 60 & 68 &          36 &          32 &        26 \\
& NMI & 76 & 86 &          46 &          38 &        45 \\
& ARI & 51 & 61 &          12 &          9 &        4 \\
& binary F1  &   83 &       87 &            75 &            70 &          64 \\
\hline
\multirow{5}{*}{\rot{CDAC}} & macro F1 &           51 &         58 &           34 &            30 &          18 \\
& ACC &         54 &       66 &          54 &          42 &        35 \\
& NMI &         74 &       86 &          67 &          51 &        61 \\
& ARI &         47 &       59 &          36 &          17 &        14 \\
& binary F1 &  82 &         83 &            74 &      72 &          68 \\
\hline
\end{tabular}
    \caption{Impact of training schemes for novel intent discovery. Models use \bertbase\ (English datasets) or \mailbert\ (Polish datasets) encoder and question input only. Individual metrics averaged over five seeds.}
    \label{tab:clustering_results_appendix}
\end{table}

\begin{table}[]
    \centering
\begin{tabular}{ll|ccc}
\hline
& \bf{Dataset} &  \rot{\tabpurchase} &  \rot{\tabdelivery} &  \rot{\tabretail} \\
\hline
\multirow{5}{*}{\emph{Q}} & macro F1  & 30 & 34 & 18 \\
& ACC &        54 & 42 & 35 \\
& NMI &        67 &    51 &        61 \\
& ARI &        36 &    17 &        14 \\
& binary F1 & 74 &    72 &        68 \\
\hline
\multirow{5}{*}{\emph{A}} & macro F1 & 27 & 23 & 12 \\
& ACC & 55 & 35 & 24 \\
& NMI & 64 & 44 & 49 \\
& ARI & 42 & 15 & 6 \\
& binary F1 & 70 & 71 & 61 \\
\hline
\multirow{5}{*}{\emph{QA concat.}} & macro F1  &  27 & 31 & 17 \\
& ACC & 59 & 45 & 41 \\
& NMI & 71 & 55 & 64 \\
& ARI & 48 & 26 & 26 \\
& binary F1  & 71 & 80 & 69 \\ 
\hline
\multirow{5}{*}{\emph{QA two head.}} & macro F1 & 38 & 32 & 19 \\
& ACC & 56 & 44 & 36 \\
& NMI & 68 & 54 & 62 \\
& ARI &  44 & 28 & 16 \\
& binary F1 &  75 & 76 & 69 \\
\hline
\multirow{5}{*}{\methodname} & macro F1 & 54 & 33 & 17 \\
& ACC & 67 & 46 & 42 \\
& NMI &         74 & 58 & 65 \\
& ARI &         52 & 32 & 27 \\
& binary F1 &  86 & 77 & 70 \\
\hline
\end{tabular}
    \caption{Impact of conversational structure for novel intent discovery.  Models use \mailbert\ initialization, CDAC training scheme, and various inputs, i.e., question \emph{Q}, answer \emph{A}, or both fields (QA) in three model variants; \emph{QA concatenation}, \emph{QA two heads}, and \methodname. Individual metrics averaged over five seeds}
    \label{tab:clustering_results_qa_appendix}
\end{table}

%% file: main.bbl
\begin{thebibliography}{27}
\expandafter\ifx\csname natexlab\endcsname\relax\def\natexlab#1{#1}\fi

\bibitem[{Angelidis and Lapata(2018)}]{angelidis-lapata-2018-summarizing}
Stefanos Angelidis and Mirella Lapata. 2018.
\newblock \href {https://doi.org/10.18653/v1/D18-1403} {Summarizing opinions:
  Aspect extraction meets sentiment prediction and they are both weakly
  supervised}.
\newblock In \emph{Proceedings of the 2018 Conference on Empirical Methods in
  Natural Language Processing}, pages 3675--3686, Brussels, Belgium.
  Association for Computational Linguistics.

\bibitem[{Bach et~al.(2018)Bach, Rodriguez, Liu, Luo, Shao, Xia, Sen, Ratner,
  Hancock, Alborzi, Kuchhal, Ré, and Malkin}]{Snorkel}
Stephen~H. Bach, Daniel Rodriguez, Yintao Liu, Chong Luo, Haidong Shao,
  Cassandra Xia, Souvik Sen, Alexander Ratner, Braden Hancock, Houman Alborzi,
  Rahul Kuchhal, Christopher Ré, and Rob Malkin. 2018.
\newblock \href {https://doi.org/10.48550/ARXIV.1812.00417} {Snorkel drybell: A
  case study in deploying weak supervision at industrial scale}.

\bibitem[{Beltagy et~al.(2019)Beltagy, Lo, and
  Cohan}]{beltagy-etal-2019-scibert}
Iz~Beltagy, Kyle Lo, and Arman Cohan. 2019.
\newblock \href {https://doi.org/10.18653/v1/D19-1371} {{S}ci{BERT}: A
  pretrained language model for scientific text}.
\newblock In \emph{Proceedings of the 2019 Conference on Empirical Methods in
  Natural Language Processing and the 9th International Joint Conference on
  Natural Language Processing (EMNLP-IJCNLP)}, pages 3615--3620, Hong Kong,
  China. Association for Computational Linguistics.

\bibitem[{Bengio et~al.(2013)Bengio, Courville, and
  Vincent}]{bengio2013representation}
Yoshua Bengio, Aaron Courville, and Pascal Vincent. 2013.
\newblock Representation learning: A review and new perspectives.
\newblock \emph{IEEE transactions on pattern analysis and machine
  intelligence}, 35(8):1798--1828.

\bibitem[{Casanueva et~al.(2020)Casanueva, Tem{\v{c}}inas, Gerz, Henderson, and
  Vuli{\'c}}]{casanueva-etal-2020-efficient}
I{\~n}igo Casanueva, Tadas Tem{\v{c}}inas, Daniela Gerz, Matthew Henderson, and
  Ivan Vuli{\'c}. 2020.
\newblock \href {https://doi.org/10.18653/v1/2020.nlp4convai-1.5} {Efficient
  intent detection with dual sentence encoders}.
\newblock In \emph{Proceedings of the 2nd Workshop on Natural Language
  Processing for Conversational AI}, pages 38--45, Online. Association for
  Computational Linguistics.

\bibitem[{{Chang} et~al.(2017){Chang}, {Wang}, {Meng}, {Xiang}, and
  {Pan}}]{DAC}
J.~{Chang}, L.~{Wang}, G.~{Meng}, S.~{Xiang}, and C.~{Pan}. 2017.
\newblock \href {https://doi.org/10.1109/ICCV.2017.626} {Deep adaptive image
  clustering}.
\newblock In \emph{2017 IEEE International Conference on Computer Vision
  (ICCV)}, pages 5880--5888.

\bibitem[{Dadas(2019)}]{polish-nlp-resources}
S{\l}awomir Dadas. 2019.
\newblock \href {https://github.com/sdadas/polish-nlp-resources/} {A repository
  of polish {NLP} resources}.
\newblock Github.

\bibitem[{Devlin et~al.(2019)Devlin, Chang, Lee, and
  Toutanova}]{devlin2019bert}
Jacob Devlin, Ming-Wei Chang, Kenton Lee, and Kristina Toutanova. 2019.
\newblock \href {http://arxiv.org/abs/1810.04805} {Bert: Pre-training of deep
  bidirectional transformers for language understanding}.

\bibitem[{Gao et~al.(2021)Gao, Arava, Hu, Mohamed, Xiao, Gao, and
  AbdelHady}]{graphire2021}
Xibin Gao, Radhika Arava, Qian Hu, Thahir Mohamed, Wei Xiao, Zheng Gao, and
  Mohamed AbdelHady. 2021.
\newblock Graphire: Novel intent discovery with pretraining on prior knowledge
  using contrastive learning.
\newblock In \emph{KDD 2021 Workshop on Pretraining: Algorithms, Architectures,
  and Applications}.

\bibitem[{Gururangan et~al.(2020)Gururangan, Marasovi{\'c}, Swayamdipta, Lo,
  Beltagy, Downey, and Smith}]{gururangan-etal-2020-dont}
Suchin Gururangan, Ana Marasovi{\'c}, Swabha Swayamdipta, Kyle Lo, Iz~Beltagy,
  Doug Downey, and Noah~A. Smith. 2020.
\newblock \href {https://doi.org/10.18653/v1/2020.acl-main.740} {Don{'}t stop
  pretraining: Adapt language models to domains and tasks}.
\newblock In \emph{Proceedings of the 58th Annual Meeting of the Association
  for Computational Linguistics}, pages 8342--8360, Online. Association for
  Computational Linguistics.

\bibitem[{Henderson et~al.(2020{\natexlab{a}})Henderson, Casanueva,
  Mrk{\v{s}}i{\'c}, Su, Wen, and Vuli{\'c}}]{henderson2020}
Matthew Henderson, I{\~n}igo Casanueva, Nikola Mrk{\v{s}}i{\'c}, Pei-Hao Su,
  Tsung-Hsien Wen, and Ivan Vuli{\'c}. 2020{\natexlab{a}}.
\newblock \href {https://doi.org/10.18653/v1/2020.findings-emnlp.196}
  {{C}onve{RT}: Efficient and accurate conversational representations from
  transformers}.
\newblock In \emph{Findings of the Association for Computational Linguistics:
  EMNLP 2020}, pages 2161--2174, Online. Association for Computational
  Linguistics.

\bibitem[{Henderson et~al.(2020{\natexlab{b}})Henderson, Casanueva,
  Mrk{\v{s}}i{\'c}, Su, Wen, and Vuli{\'c}}]{henderson-etal-2020-convert}
Matthew Henderson, I{\~n}igo Casanueva, Nikola Mrk{\v{s}}i{\'c}, Pei-Hao Su,
  Tsung-Hsien Wen, and Ivan Vuli{\'c}. 2020{\natexlab{b}}.
\newblock \href {https://doi.org/10.18653/v1/2020.findings-emnlp.196}
  {{C}onve{RT}: Efficient and accurate conversational representations from
  transformers}.
\newblock In \emph{Findings of the Association for Computational Linguistics:
  EMNLP 2020}, pages 2161--2174, Online. Association for Computational
  Linguistics.

\bibitem[{Huang et~al.(2019)Huang, Altosaar, and Ranganath}]{clinicalbert}
Kexin Huang, Jaan Altosaar, and Rajesh Ranganath. 2019.
\newblock Clinicalbert: Modeling clinical notes and predicting hospital
  readmission.
\newblock \emph{arXiv:1904.05342}.

\bibitem[{Larson et~al.(2019)Larson, Mahendran, Peper, Clarke, Lee, Hill,
  Kummerfeld, Leach, Laurenzano, Tang, and Mars}]{Clinc150}
Stefan Larson, Anish Mahendran, Joseph~J. Peper, Christopher Clarke, Andrew
  Lee, Parker Hill, Jonathan~K. Kummerfeld, Kevin Leach, Michael~A. Laurenzano,
  Lingjia Tang, and Jason Mars. 2019.
\newblock \href {http://arxiv.org/abs/1909.02027} {An evaluation dataset for
  intent classification and out-of-scope prediction}.
\newblock \emph{CoRR}, abs/1909.02027.

\bibitem[{Lin et~al.(2020)Lin, Xu, and Zhang}]{CDAC}
Ting{-}En Lin, Hua Xu, and Hanlei Zhang. 2020.
\newblock \href {https://aaai.org/ojs/index.php/AAAI/article/view/6353}
  {Discovering new intents via constrained deep adaptive clustering with
  cluster refinement}.
\newblock In \emph{The Thirty-Fourth {AAAI} Conference on Artificial
  Intelligence, {AAAI} 2020, The Thirty-Second Innovative Applications of
  Artificial Intelligence Conference, {IAAI} 2020, The Tenth {AAAI} Symposium
  on Educational Advances in Artificial Intelligence, {EAAI} 2020, New York,
  NY, USA, February 7-12, 2020}, pages 8360--8367. {AAAI} Press.

\bibitem[{Lloyd(1982)}]{Lloyd1982LeastSQ}
Stuart~P. Lloyd. 1982.
\newblock Least squares quantization in pcm.
\newblock \emph{IEEE Trans. Inf. Theory}, 28:129--136.

\bibitem[{Mroczkowski et~al.(2021)Mroczkowski, Rybak, Wr{\'o}blewska, and
  Gawlik}]{mroczkowski-etal-2021-herbert}
Robert Mroczkowski, Piotr Rybak, Alina Wr{\'o}blewska, and Ireneusz Gawlik.
  2021.
\newblock \href {https://aclanthology.org/2021.bsnlp-1.1} {{H}er{BERT}:
  Efficiently pretrained transformer-based language model for {P}olish}.
\newblock In \emph{Proceedings of the 8th Workshop on Balto-Slavic Natural
  Language Processing}, pages 1--10, Kiyv, Ukraine. Association for
  Computational Linguistics.

\bibitem[{Reimers and Gurevych(2019)}]{reimers-gurevych-2019-sentence}
Nils Reimers and Iryna Gurevych. 2019.
\newblock \href {https://doi.org/10.18653/v1/D19-1410} {Sentence-{BERT}:
  Sentence embeddings using {S}iamese {BERT}-networks}.
\newblock In \emph{Proceedings of the 2019 Conference on Empirical Methods in
  Natural Language Processing and the 9th International Joint Conference on
  Natural Language Processing (EMNLP-IJCNLP)}, pages 3982--3992, Hong Kong,
  China. Association for Computational Linguistics.

\bibitem[{Reimers and Gurevych(2020)}]{reimers-gurevych-2020-making}
Nils Reimers and Iryna Gurevych. 2020.
\newblock \href {https://doi.org/10.18653/v1/2020.emnlp-main.365} {Making
  monolingual sentence embeddings multilingual using knowledge distillation}.
\newblock In \emph{Proceedings of the 2020 Conference on Empirical Methods in
  Natural Language Processing (EMNLP)}, pages 4512--4525, Online. Association
  for Computational Linguistics.

\bibitem[{Rybak et~al.(2020)Rybak, Mroczkowski, Tracz, and
  Gawlik}]{rybak-etal-2020-klej}
Piotr Rybak, Robert Mroczkowski, Janusz Tracz, and Ireneusz Gawlik. 2020.
\newblock \href {https://doi.org/10.18653/v1/2020.acl-main.111} {{KLEJ}:
  Comprehensive benchmark for {P}olish language understanding}.
\newblock In \emph{Proceedings of the 58th Annual Meeting of the Association
  for Computational Linguistics}, pages 1191--1201, Online. Association for
  Computational Linguistics.

\bibitem[{Song et~al.(2020)Song, Tan, Qin, Lu, and Liu}]{NEURIPS2020_c3a690be}
Kaitao Song, Xu~Tan, Tao Qin, Jianfeng Lu, and Tie-Yan Liu. 2020.
\newblock \href
  {https://proceedings.neurips.cc/paper/2020/file/c3a690be93aa602ee2dc0ccab5b7b67e-Paper.pdf}
  {Mpnet: Masked and permuted pre-training for language understanding}.
\newblock In \emph{Advances in Neural Information Processing Systems},
  volume~33, pages 16857--16867. Curran Associates, Inc.

\bibitem[{Tracz et~al.(2020)Tracz, W{\'o}jcik, Jasinska-Kobus, Belluzzo,
  Mroczkowski, and Gawlik}]{tracz-etal-2020-bert}
Janusz Tracz, Piotr~Iwo W{\'o}jcik, Kalina Jasinska-Kobus, Riccardo Belluzzo,
  Robert Mroczkowski, and Ireneusz Gawlik. 2020.
\newblock \href {https://aclanthology.org/2020.ecomnlp-1.7} {{BERT}-based
  similarity learning for product matching}.
\newblock In \emph{Proceedings of Workshop on Natural Language Processing in
  E-Commerce}, pages 66--75, Barcelona, Spain. Association for Computational
  Linguistics.

\bibitem[{Vedula et~al.(2022)Vedula, Gupta, Alok, Sridhar, and
  Ananthakrishnan}]{advin2022}
Nikhita Vedula, Rahul Gupta, Aman Alok, Mukund Sridhar, and Shankar
  Ananthakrishnan. 2022.
\newblock \href {https://doi.org/10.1109/ICASSP43922.2022.9746672} {Advin:
  Automatically discovering novel domains and intents from user text
  utterances}.
\newblock In \emph{ICASSP 2022 - 2022 IEEE International Conference on
  Acoustics, Speech and Signal Processing (ICASSP)}, pages 7627--7631.

\bibitem[{Wang et~al.(2018)Wang, Wang, Zhou, Ji, Gong, Zhou, Li, and
  Liu}]{8578650}
Hao Wang, Yitong Wang, Zheng Zhou, Xing Ji, Dihong Gong, Jingchao Zhou, Zhifeng
  Li, and Wei Liu. 2018.
\newblock \href {https://doi.org/10.1109/CVPR.2018.00552} {Cosface: Large
  margin cosine loss for deep face recognition}.
\newblock In \emph{2018 IEEE/CVF Conference on Computer Vision and Pattern
  Recognition}, pages 5265--5274.

\bibitem[{Wang et~al.(2020)Wang, Bi, Yan, Wu, Xia, Bao, Peng, and
  Si}]{structbert}
Wei Wang, Bin Bi, Ming Yan, Chen Wu, Jiangnan Xia, Zuyi Bao, Liwei Peng, and
  Luo Si. 2020.
\newblock \href {https://openreview.net/forum?id=BJgQ4lSFPH} {Structbert:
  Incorporating language structures into pre-training for deep language
  understanding}.
\newblock In \emph{8th International Conference on Learning Representations,
  {ICLR} 2020, Addis Ababa, Ethiopia, April 26-30, 2020}. OpenReview.net.

\bibitem[{Zhang et~al.(2020)Zhang, Xu, Lin, and Lv}]{zhang_dac2020}
Hanlei Zhang, Hua Xu, Ting{-}En Lin, and Rui Lv. 2020.
\newblock \href {http://arxiv.org/abs/2012.08987} {Discovering new intents with
  deep aligned clustering}.
\newblock \emph{CoRR}, abs/2012.08987.

\bibitem[{Zheng and Lapata(2019)}]{zheng-lapata-2019-sentence}
Hao Zheng and Mirella Lapata. 2019.
\newblock \href {https://doi.org/10.18653/v1/P19-1628} {Sentence centrality
  revisited for unsupervised summarization}.
\newblock In \emph{Proceedings of the 57th Annual Meeting of the Association
  for Computational Linguistics}, pages 6236--6247, Florence, Italy.
  Association for Computational Linguistics.

\end{thebibliography}
